\RequirePackage{fix-cm}
\documentclass[twocolumn]{svjour3}          
\smartqed  
\usepackage{graphicx}
 \usepackage{mathptmx}      
\usepackage[linesnumbered,vlined,ruled]{algorithm2e}
\usepackage{amsmath, amssymb, bm}
\usepackage{epsfig}
\usepackage{graphics}
\usepackage{grffile}
\usepackage[utf8]{inputenc}
\usepackage{listings}
\usepackage{multirow}
\usepackage[figuresright]{rotating}
\usepackage{sidecap}
\usepackage{syntax}
\usepackage{url}
\usepackage{verbatim}
\usepackage{xspace}
\usepackage{tikz}
\usepackage[misc,geometry]{ifsym}
\newcommand{\fig}[1]{Fig.~\ref{#1}}

\newcommand{\sect}[1]{Sec.~\ref{#1}}

\newcommand{\etal}[0]{{\em et al.~}}
\newcommand{\eg}[0]{{\em e.g.,~}}
\newcommand{\ie}[0]{{\em i.e.,~}}

\journalname{International Journal of Social Robotics}
\begin{document}
\sloppy
\title{iRoPro: An interactive Robot Programming Framework
}

\titlerunning{iRoPro: An interactive Robot Programming Framework}        

\author{Ying Siu Liang \and Damien Pellier \and
	Humbert Fiorino \and Sylvie Pesty
}

\authorrunning{Ying Siu Liang et al.} 

\institute{Ying Siu Liang \Letter \and Damien Pellier \and Humbert Fiorino \and Sylvie Pesty  \at
              Universit\'{e} Grenoble Alpes, LIG, 38000 Grenoble, France \\
              \email{\{liangyi, pellierd, fiorinhu, pestys\}@univ-grenoble-alpes.fr}           
}

\date{Received: date / Accepted: date}

\maketitle

\begin{abstract}
The great diversity of end-user tasks ranging from manufacturing environments to personal homes makes pre-programming robots for general purpose applications extremely challenging.
In fact, teaching robots new actions from scratch that can be reused for previously unseen tasks remains a difficult challenge and is generally left up to robotics experts.
In this work, we present iRoPro, an interactive Robot Programming framework that allows end-users with little to no technical background to teach a robot new reusable actions.
We combine Programming by Demonstration and Automated Planning techniques to allow the user to construct the robot's knowledge base by teaching new actions by kinesthetic demonstration.
The actions are generalised and reused with a task planner to solve previously unseen problems defined by the user.
We implement iRoPro as an end-to-end system on a Baxter Research Robot to simultaneously teach low- and high-level actions by demonstration that the user can customise via a Graphical User Interface to adapt to their specific use case.
To evaluate the feasibility of our approach, we first conducted pre-design experiments to better understand the user's adoption of involved concepts and the proposed robot programming process.
We compare results with post-design experiments, where we conducted a user study to validate the usability of our approach with real end-users.
Overall, we showed that users with different programming levels and educational backgrounds can easily learn and use iRoPro and its robot programming process.

\end{abstract}

\section{Introduction} \label{intro}
Despite the ongoing advances in Robotics and A.I., it is extremely challenging to pre-program robots for specific end-user applications.
Instead of developing robots for domain-specific tasks, a more flexible solution is to have them learn new actions directly from end-users who can customise the robot for their specific application.
Programming by Demonstration (PbD) \cite{billard2008robot} is a popular approach for end-users to teach robots actions in an intuitive way by taking demonstrations as input and inferring a policy for the task.
However, PbD solutions usually require users to teach robots an action sequence to achieve a certain goal.
If the goal changes, the user has to teach the robot a new sequence.\\

Consider the Tower of Hanoi, a puzzle consisting of three pegs and a number of differently-sized disks, stacked on one peg in descending order, with the largest peg at the bottom.
The goal is to move the entire stack from one peg to another, by moving one disk at a time and only to a larger disk or an empty peg.
The solution is different depending on the given number of disks.
If we want to teach a robot to solve this problem, it would be infeasible to demonstrate the solution each time.
A more efficient approach would be to teach the robot the primitive action of moving a disk, associate rules or \textit{conditions} to this action (\eg smaller disks can only be placed on top of larger ones), and have the robot generate an optimal solution.\\

Different approaches to generate robot actions have been proposed \cite{8523933}. The {\it reactive} approach is characterised by independent and concurrent basic behaviours that generate the robot global behaviour. The robot sensors are the input of the basic behaviours that, in turn, activate/inhibit the robot effectors. This approach is effective but predicting the robot global behaviour is difficult since no explicit action model is used. The {\it deliberative} approach consists of three steps: perception of the world state, plan making and action execution in order to change the world state. Plan making also known as Automated Planning \cite{ghallab2004automated} is based on an explicit action model encoded in a symbolic planning language. This action model is used by a task planner to compute action sequences achieving the expected robot behaviours.

Our research argues for teaching robots primitive actions, instead of entire action sequences, and delegating the logical reasoning process of finding a solution to a task planner.
To this end, we present iRoPro, an interactive Robot Programming framework that allows efficient programming of both \textit{how} an action is performed (low-level action representation) and \textit{when} it can be applied (high-level), while generalising both aspects to new scenarios.
The low-level representation allows the robot to execute the motion trajectory, while the high-level representation allows it to be used with a task planner.

We implement the framework on a Baxter robot and develop an intuitive graphical interface that allows users to teach new actions by demonstration and customise them to be reused for new problems that can be solved with a task planner. 
The developed end-to-end system involves solutions in perception (\eg object identification), motion planning (\eg manipulation, navigation, safety), cognitive robotics (\eg action learning, task planning) and human–robot interaction (\eg multi-modal interaction and teacher feedback).
Even though task planners are generally used by domain-experts, we show that users with little to no programming experience can easily learn and use their main concepts.
We conduct pre-design experiments, where we simulate the framework with the Wizard-of-Oz technique to evaluate the usability of the programming process.
We compare these results with post-design experiments and empirically investigate our system's usability with a user study (N=21) where real end-users programmed a Baxter robot directly with our end-to-end system.

In \sect{sec:background} we give a brief overview of PbD and Automated Planning, the two underlying techniques of the framework.
In \sect{sec:approach} we present iRoPro, the interative Robot Programming framework and its main components.
Then in \sect{sec:implementation} we provide details of the system implemented on a Baxter robot and the user programming process.
\sect{sec:exp-eval} presents the experimental evaluation of our approach, where we compare pre-design experiments with post-design experiments.
Finally, in \sect{sec:conclusion}, we conclude by discussing limitations and possible extensions to further increase the system's generalisability.

\section{Background}\label{sec:background}
Traditional robot programming processes have task-specific definitions, are generally robot-dependent and require programming expertise.
In the past few decades, different techniques have been developed to facilitate the robot programming process.
Biggs \etal \cite{Biggs2003} defined two main categories, manual and automatic programming, distinguishing between systems where users can or cannot directly control the robot's executed behaviour.

In manual programming, the user encodes the robot's behaviour via text-based systems using procedural languages such as python, C++, java or graphical systems that use a graph, flow-chart or diagram \cite{lego2003,fraser2013blockly,majed2014learn}.
For automatic programming techniques, robots generate their behaviour from data provided as input to the system.
We differentiate between Deep Learning (DL) \cite{schmidhuber2015deep}, where the robot is provided a large amount of labelled or unlabelled data, Reinforcement Learning \cite{kaelbling1996reinforcement,gosavi2009reinforcement}, where the robot gathers the data by exploring the environment, and Programming by Demonstration (PbD) \cite{billard2008robot,argall2009survey}, where the robot learns from example demonstrations provided by a human teacher.
While DL approaches allow the robot to learn skills autonomously, they often require programming and domain experts to prepare the system input (\eg label or preprocess data, define policy and reward functions).
Similarly, reinforcement learning generally requires long training times to gather enough data to learn a skill.

In contrast, PbD provides a more intuitive low-effort solution, where the teacher's main task involves providing demonstrations to the robot.
Since PbD solutions allow the robot to learn from a sparse set of examples, the data and time required to learn a skill is moderately low.
Thus, our work uses PbD to allow end-users to program robots.
In the following sections we will give a brief overview of PbD and Automated Planning, the two main concepts used in our proposed framework.

\subsection{Programming by Demonstration}\label{subsec:RPbD}
PbD, also referred to as \textit{Learning from Demonstration}, is an end-user programming technique for teaching a robot new skills by demonstrating them, without writing code \cite{billard2008robot}.
It has become a central topic in research areas, with the aim to move from purely pre-programmed robots to flexible user-based interfaces for training robots.
PbD is traditionally used to learn low-level actions from \textit{trajectory} demonstrations using Gaussian Mixture Models \cite{billard2008robot,calinon2007incremental} or Dynamic Movement Primitives \cite{pastor2009learning}.
They can also be learned from \textit{keyframe-based} demonstrations (kfPbD), where the user kinesthetically manipulates the robot's arm to record a series of end-effector poses, referred to as \textit{keyframes} \cite{akgun2012keyframe}.
While demonstrations can be provided in different ways (\eg by observating a human teacher), users prefer to control the robot directly \cite{suay2012practical}.
In kfPbD, actions are represented as a sparse sequence of gripper states (open/close) and end-effector poses relative to perceived objects or to the robot's coordinate frame.
Alexandrova \etal \cite{alexandrova2014robot} implemented an end-user robot programming system to teach generalisable actions from a single demonstration where keyframes are automatically inferred and actions can be modified retrospectively via a graphical interface.

\begin{figure}
	\centering
	\includegraphics[width=\linewidth]{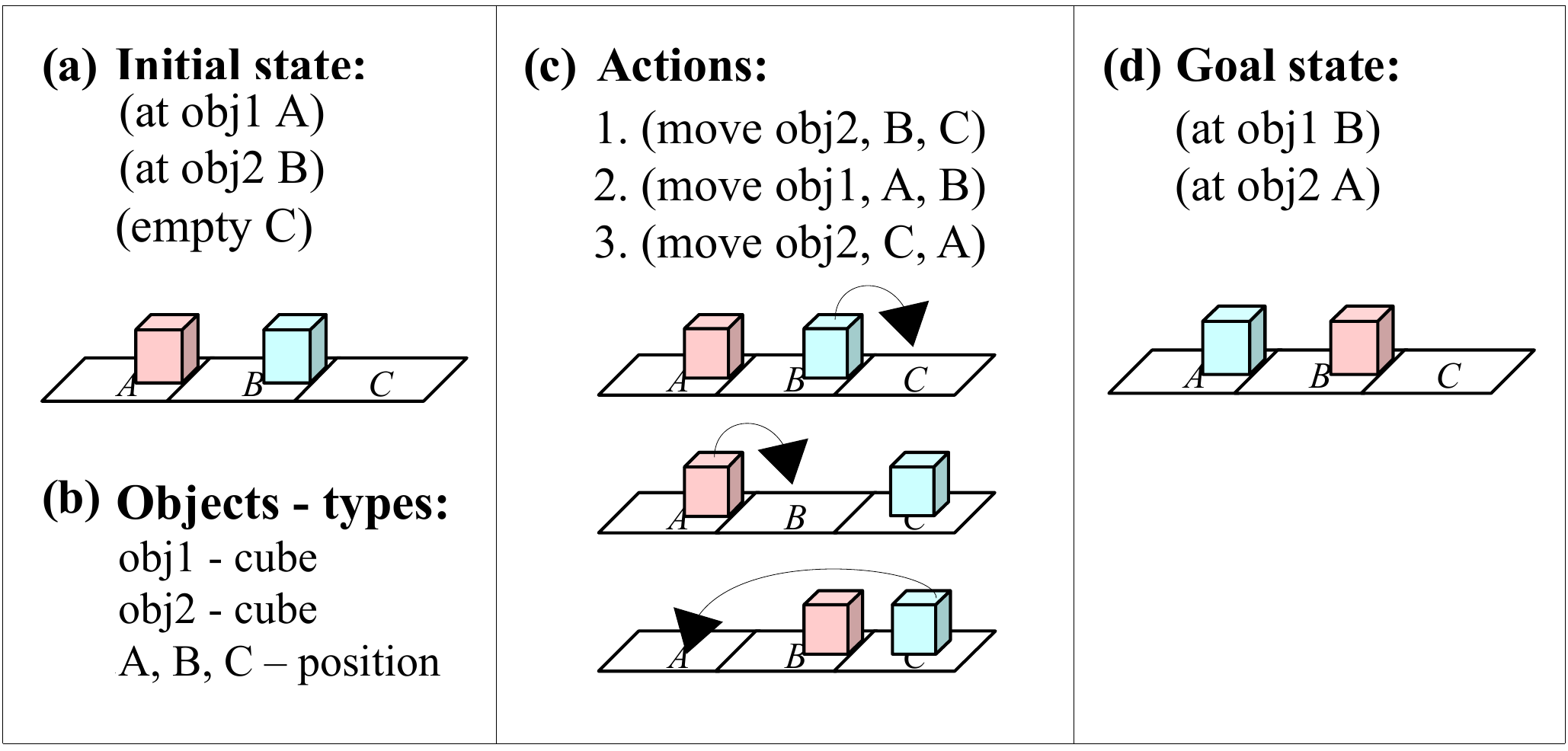}
	\caption{Definition of a planning problem (a) properties describing the initial world state (b) object names and their types (c) instantiated actions (d) properties describing the goal \cite{liang2017b}.}
	\label{fig:planning-permutation}
\end{figure}

\begin{figure}
	\centering
	\includegraphics[width=\linewidth]{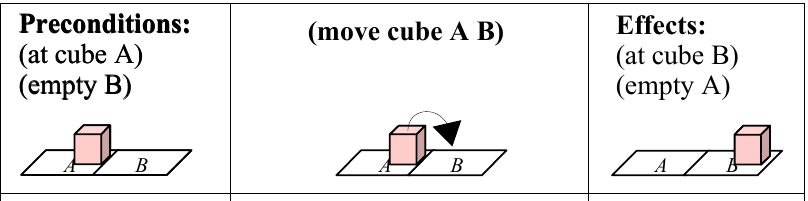}
	\caption{Action model representation to move a cube from position A to B in terms of preconditions and effects \cite{liang2017b}.}
	\label{fig:actionmodel-exp2}
\end{figure}

\subsection{Automated Planning}\label{subsec:AP}
\begin{figure}
	\centering
	\includegraphics[width=\linewidth]{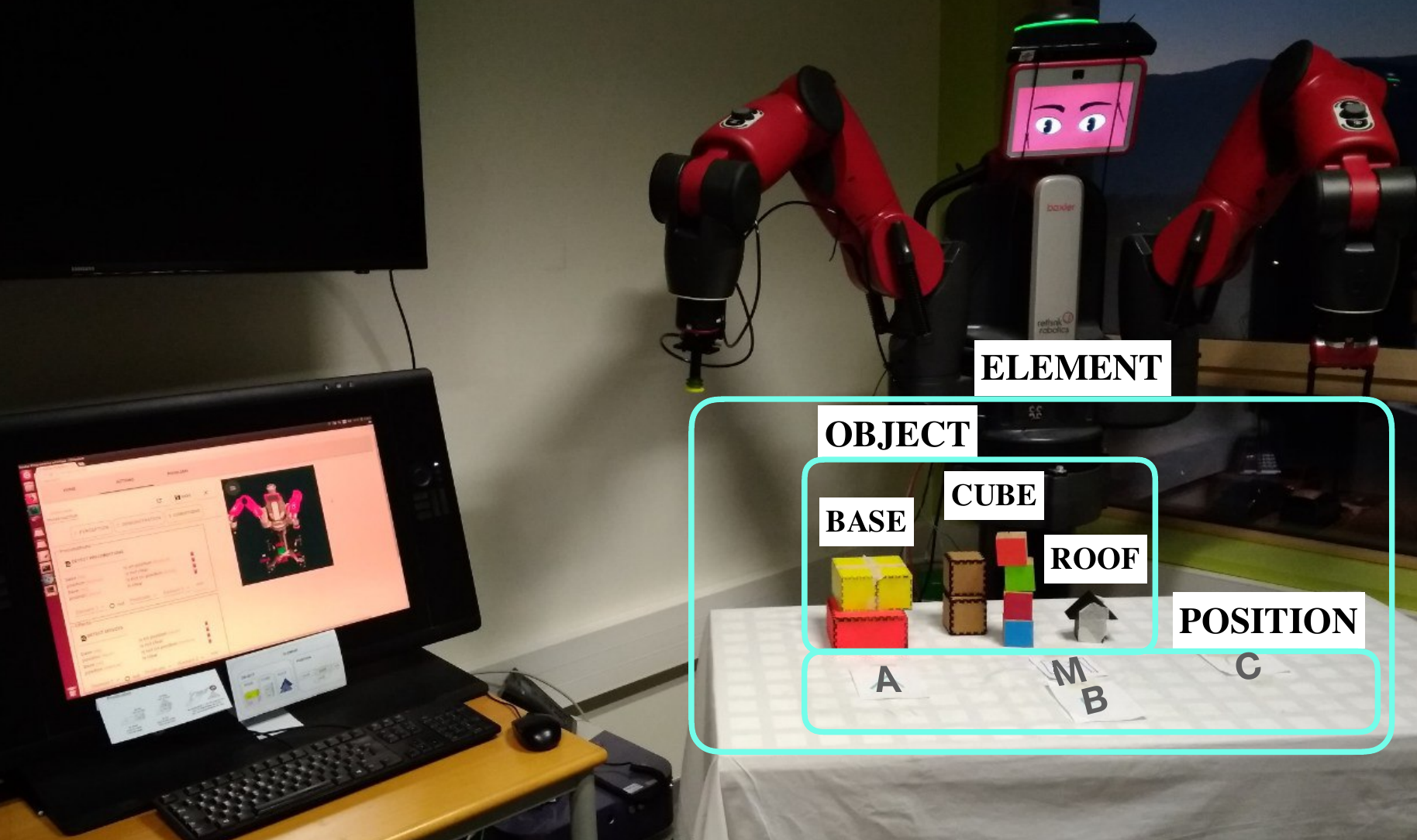}
	\caption{Experimental setup of the user study. Users programmed the Baxter robot via a graphical interface in order to manipulate objects (shown with their type hierarchies) in the task domain \cite{liang2019d}. }
	\label{fig:dispositif}
\end{figure}

Automated Planning, also known as \textit{AI Planning}, is a research field that focuses on the development of task planners consisting of efficient search algorithms to generate solutions to problems \cite{ghallab2004automated}.
Given a planning \textit{domain}, \ie a description of the state of the world and a set of actions, and a planning \textit{problem}, \ie an initial state (\fig{fig:planning-permutation} (a)) and a goal (\fig{fig:planning-permutation} (d)), the task planner generates a sequence of actions (\fig{fig:planning-permutation} (c)), which guarantees the transition from initial states to goal.
To allow a correct transition between different world states, high-level actions are defined in terms of preconditions and effects, which represent states before and after the action execution respectively (\fig{fig:actionmodel-exp2}).
An action is represented as a tuple $a = (\text{param}(a), \text{pre}(a),$ $\text{eff}(a))$, whose elements are:
\begin{itemize}
	\item $\text{param}(a)$: set of parameters that $a$ applies to
	\item $\text{pre}(a)$: set of predicates that must be true to apply $a$
	\item $\text{eff}(a)^{-}$: set of predicates that are false after applying $a$
	\item $\text{eff}(a)^{+}$: set of predicates that are true after applying $a$
\end{itemize}
where $\text{eff}(a) = \text{eff}(a)^{-} \cup \text{eff}(a)^{+}$.
Action parameters are associated with a \textit{type} (\fig{fig:planning-permutation} (b)) and a potential type hierarchy.
For example, a type hierarchy, consisting of a general type ELEMENT, divided into POSITION and OBJECT, which further divides into BASE, CUBE, and ROOF (\fig{fig:dispositif}).
Predicates are defined in first-order logic and used to describe world states and relations between their elements (\eg \texttt{(at obj1 A)}).
Planning algorithms use a symbolic planning language as their standard encoding language, such as STRIPS \cite{fikes1971strips} or PDDL \cite{ghallab2004automated}.
An example of a planning domain in PDDL can be seen in \fig{fig:pddl}.

The Tower of Hanoi problem could be defined in terms of a planning problem, where the domain consists of 3 pegs and a number of disks, and the action is defined as moving a disk from one peg to another, with associated rules as preconditions and effects.
A planner can then be used to generate a solution to the problem for any number of disks.

From now on, the word "action" will refer to an action as defined in automated planning. It is worth noting that the predicates used in the planning domains are freely defined by the persons encoding those domains. However, creating a standard for knowledge representation and reasoning in autonomous robotics is a basic requirement to allow the interoperability of the robotic systems, the communication between robots and humans etc. based on a common vocabulary along with clear and concise definitions. The IEEE Standard Association's Robotics and Automation Society recognized this need, and a set of ontologies have been developped \cite{8172300}.

\begin{figure}
\begin{verbatim}
(define (domain iRoPro)
 (:requirements :strips :typing)
 (:types
    element
    position - element
    object - element
    cube - object
    base - object
    roof - object )
 (:predicates
    (clear ?e - element)
    (thin ?o - object)
    (flat ?e - element)
    (on ?o - object ?e - element)
    (stackable ?o - object ?e - element)
 (:action move
  :parameters (?o - object ?A - position ?B - position)
  :precondition (and (on ?o ?A) (clear ?o) (clear ?B)
  :effect (and (on ?o ?B) (clear ?A)
                      not(on ?o ?A) not(clear ?B))
)
\end{verbatim}
    \caption{Example of a planning domain in PDDL.}
	\label{fig:pddl}
\end{figure}

\begin{figure*}
	\centering
	\includegraphics[width=0.7\textwidth]{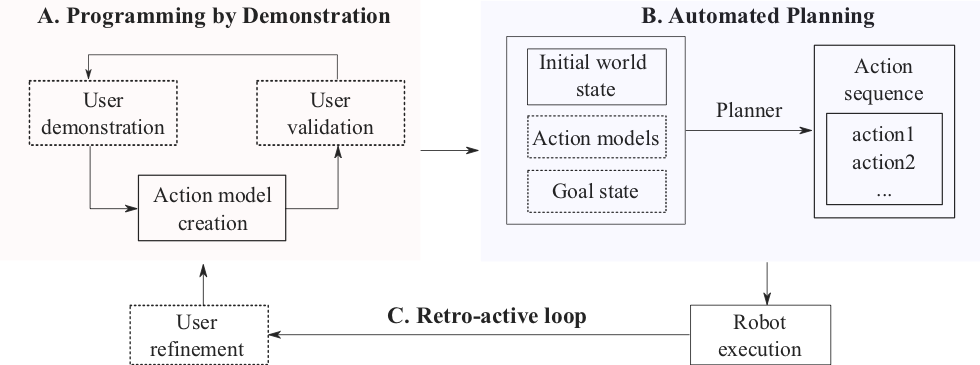}
	\caption{Overview of the main components of iRoPro: A. the user teaches primitive actions by demonstration B. the robot reuses these with a task planner to generate an action sequence to achieve a goal
		C. After observing the robot execution, the user can refine the taught action models \cite{liang2017b}.}
	\label{fig:framework}
\end{figure*}

\section{Related Work} \label{sec:relatedwork}
End-user robot programming has been addressed previously for industrial robots to be programmed by non-robotics domain experts, where users specify and modify existing plans for robots to adapt to new scenarios \cite{paxton2017costar,perzylo2016intuitive,stenmark2017simplified}.
For example, Paxton \etal \cite{paxton2017costar} use Behaviour Trees to represent task plans that are explicitly defined by the user and can be modified to adapt to new tasks.
In our work we argue for the use of task planners to automatically generate plans for new scenarios, rather than have the user manually modify them.

Previous work has addressed knowledge engineering tools for constructing planning domains but usually require PDDL experts (PDDL Studio \cite{plch2012inspect}), or common knowledge in software engineering (GIPO \cite{simpson2007planning}, itSIMPLE \cite{vaquero2013itsimple}).
There has been previous work on integrating task planning with robotic systems \cite{cashmore2015rosplan,kuhner2018closed}, learning high-level actions through natural language instructions \cite{she2014teaching} or learning preconditions and effects of actions to be used in planning \cite{jetchev2013learning,konidaris2018fromSkills,ugur2015bottom}.
However, in all of these cases, the robot is provided with a fixed set of low-level motor skills.
In our approach, we do not provide the robot with any predefined actions but enable users to teach both low- and high-level actions from scratch.

PbD has been commonly applied to allow end-users to teach robots new actions by demonstration.
Alexandrova \etal \cite{alexandrova2014robot} created an end-user programming framework with an interactive action visualisation allowing the user to teach new actions from single demonstrations but do not reuse them with a task planner.
Most closely related to our approach is the work by Abdo \etal \cite{abdo2013learning} where manipulation actions are learned from kinesthetic demonstrations and reused with task planners.
However, the approach requires 5-10 demonstrations to learn action conditions which becomes tedious and impractical if several actions need to be taught.
In this work we argue for having the user act as the expert by letting them correct inferred action conditions, thus allowing a new action to be learned from a single demonstration.
We further provide a graphical interface that allows users to create new actions and address previously unseen problems that can be solved with task planners.

\section{iRoPro - interactive Robot Programming} \label{sec:approach}
In our previous work \cite{liang2017a} we proposed iRoPro, an interactive Robot Programming framework that allows end-users to teach robots new actions that can be reused with task planners.
The framework consists of the following three aspects (\fig{fig:framework}):
\begin{enumerate}
	\item[A.]{Programming by Demonstration: The user \textit{teaches} the robot primitive actions by demonstration. The robot creates an action model that the user can modify and validate.}
	\item[B.]{Automated Planning: The user defines a new planning problem with a goal to achieve. The robot \textit{reuses} the taught actions with a planner to generate solutions for new problems.}
	\item[C.]{Retro-active Loop: The user observes the robot execution and \textit{refines} taught actions via the graphical interface.}
\end{enumerate}
The user is provided with a GUI that abstracts from the underlying modeling language used for Automated Planning.
For each step, the user interacts with the GUI to navigate between the components to teach new actions by demonstration, modify inferred action conditions, define new planning problems for the robot to solve and execute generated plans.
In the following sections, we give a brief description of each component.
We refer the reader to our previous work \cite{liang2017a,liang2017b} for more details.

\subsection{Programming by Demonstration: teach actions}
Teaching primitive actions consists of learning both \textit{how} and \textit{when} an action should be applied, \ie learning the low-level action trajectory as in PbD and the high-level representation with preconditions and effects as used in Automated Planning.
We consider an action that consists of both low- and high-level representations an \textit{action model}.
The high-level representation can be entered directly by the user or inferred from observing the world state before and after the action demonstration.
In our work, we first infer the preconditions and effects which the user can subsequently modify on the GUI.
We rely on the user's logical reasoning and understanding of what they want to teach the robot and allow them to directly program and correct inferred action models.
Thus, the robot can learn a new action from a single demonstration with the user acting as the expert to correct inferred conditions.
The user validates the learned action model or provides additional demonstrations to refine the low- or high-level representations.
The user repeats this process and creates an action model for each primitive action (see \fig{fig:framework}A).

\subsection{Automated Planning: reuse actions}\label{sec:AP}
In the PbD step, the robot learned action models that include the high-level representation used in Automated Planning.
The Automated Planning step consists of creating a planning problem and defining a goal for which the integrated task planner can generate a solution (\fig{fig:framework}B).
Given a description of a planning \textit{domain}, we can define a planning \textit{problem} with an initial state and a desired goal state to achieve.
Depending on the robot architecture and perception system, a partial PDDL domain can be predefined in the system that includes a set of object types and predicates that the robot can recognise.
The action models created in the PbD step complete the partial PDDL domain.
Similar as before, the robot can infer initial world states for the planning problem and the user can modify and correct them via the GUI.
Given the user-defined goal, the task planner generates a plan consisting of an ordered action sequence for the robot to execute.
The user can verify the generated plan and have the robot execute it in real life.
If no plan is generated or if the plan seems incorrect, the user can modify the taught action models, as well as the initial and goal states and relaunch the planner.

\begin{figure*}
	\centering
	\includegraphics[width=0.99\textwidth]{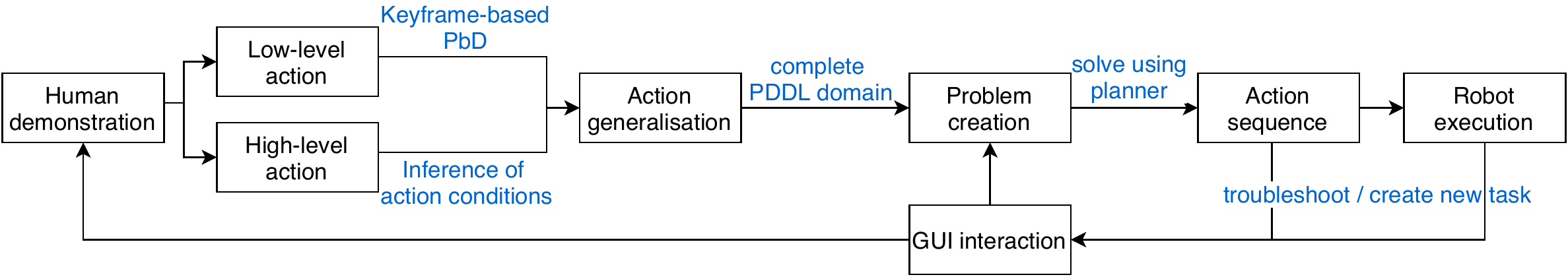}
	\caption{Overview of iRoPro that allows users to teach low- and high-level actions by demonstration. The user interacts with the GUI to run the demonstration, modify inferred action conditions, create new planning problems for the robot to solve and execute.}
	\label{fig:overview}
\end{figure*}
\subsection{Retro-active Loop: refine actions}
The retro-active loop allows the user to revisit and correct created action models (see \fig{fig:framework}C).
It is likely that the initially generated plan would not achieve the specified goal, especially if the context of the planning problem is different to that of the initial demonstration (\eg different object types or positions).
Programmed action models can be generalised and reused, especially if the low-level action remains the same.
Instead of creating new action models for each new problem, the user can simply modify the action parameters, preconditions or effects.
This minimises the user's programming effort and the number of demonstrations required.
Thus, the application to a new context is an important step to test the generalisability of action models.

There are several possible causes why the planner might generate incorrect or non-existent solutions:
\begin{itemize}
	\item \textbf{Action parameters:} this restricts or generalises the application of the action as they dictate what types an action can be applied to. An action is not considered by the planner, if the types do not match those in the initial state of the planning problem (\eg pick-and-place was only defined for cube objects but not for other types).
	\item \textbf{Preconditions:} similar to action parameters, they define when an action can be applied, but in terms of predicates describing the initial world state. All stated preconditions must hold in a world state in order to apply the action (\eg pick-and-place of an object only if it is clear).
	\item \textbf{Effects:} they define how the world state is updated after the action execution and help the robot to keep track of changes. If they are not defined correctly, there can be a mismatch of the robot's assumed world state and the actual world state (\eg a position is still considered free when it is occupied).
	\item \textbf{Initial states:} they describe the existing world state to the robot. If the initial states are incorrect or missing the planner may consider certain actions as invalid or not find a plan to the goal (\eg an object is not mentioned in the initial states at all).
	\item \textbf{Goal:} the user defines the set of predicates for the robot to achieve. This  should not include intermediate steps to achieve the goal nor contradicting states (\eg `object is on A' and `A is clear' are both stated as goal states).
\end{itemize}

Knowledge engineering tools can facilitate this process of modifying action models.
They often provide useful functionalities for dynamic testing, model checking and visualisation \cite{simpson2007planning}, but most tools require expertise in Automated Planning or Software Engineering.
In our work we argue that the proposed robot programming process does not require this expertise and can be learned easily by users with different educational backgrounds.

\section{System Implementation} \label{sec:implementation}
We implemented iRoPro on a Baxter robot with two arms (one claw and one suction gripper), both with 7-DoF and a load capacity of 2.2kg.
For the object perception we mounted a Kinect Xbox 360 depth camera on the robot.
We developed a user interface as a web application that can be accessed via a browser on a PC, tablet or smartphone.
The source code for iRoPro is developed in ROS \cite{quigley2009ros} and available online\footnote{\url{https://github.com/ysl208/iRoPro/tree/cond}}.
The action is learned by demonstration using the open-source system Rapid PbD\footnote{\url{https://github.com/jstnhuang/rapid_pbd}}.
The integration of the task planner is implemented using the ROS package PDDL planner\footnote{\url{http://docs.ros.org/indigo/api/pddl_planner}}.
In our implementation, we define \textit{landmarks} as either predefined table positions or objects that are detected from Kinect point cloud clusters using an open-source tabletop segmentation library\footnote{\url{https://github.com/jstnhuang/surface_perception}}.
An object $o = (x,y,z, width, length, height)$ is represented by its detected location and bounding box, which are used to infer its type and related predicates.
The user completes the partial PDDL domain via the GUI by creating new action models by demonstration.
Then they create planning problems that can be solved with the integrated task planner.
Figure \ref{fig:overview} shows an overview of the programming process.\footnote{Video can be seen at \url{https://youtu.be/NgaTPG8dZwg}}
In the following sections we will give a brief overview of the system implementation.
Further details can be found in our previous work \cite{liang2019d}.

\subsection{Low-level Action Representation}
\label{sec:lowlevel}
We represent low-level actions as proposed in previous work using keyframe-based PbD \cite{alexandrova2014robot}, where the action is represented as a sparse sequence of gripper states (open/close) and end-effector poses relative to perceived objects or to the robot's coordinate frame.
During the action demonstration, the user guides the robot arm using kinesthetic manipulation and saves poses that they find relevant for the action.
For example, the pick-and-place action of an object to a marked position could be represented as poses relative to the object (for the pick action), poses relative to the target position (for the place action), and corresponding open/close gripper states.
Action executions are performed by first detecting the landmarks in the environment, calculating the end-effector poses relative to the observed landmarks, and interpolating between the poses.
While these actions can be learned from multiple demonstrations \cite{niekum2012learning}, we take the approach that only requires a single demonstration by heuristically assigning poses and letting the user correct them if needed \cite{alexandrova2014robot}.
Thus, the first demonstrated action is already an executable action.
The user can teach multiple manipulation actions and discriminate between them by associating different conditions that specify \textit{when} the robot should use them (\eg different conditions for actions using claw or suction grippers).

\subsection{High-level Action Representation}
\label{sec:highlevel}
We implemented a partial PDDL domain with predefined types and predicates that the robot can automatically detect using its sensors.
We defined five predicates commonly used for object manipulation tasks and included two (flat and thin) to describe further object properties:
\begin{itemize}
	\item \textit{ELEMENT is clear}: an element has nothing on top of it
	\item \textit{OBJECT is on ELEMENT}: an object is on an element
	\item \textit{OBJECT is stackable on ELEMENT}: an object can be placed on an element
	\item \textit{OBJECT is flat}: an object has a flat top
	\item \textit{OBJECT is thin}: an object can be grasped with the claw gripper
\end{itemize}
In our definition, CUBE and BASE objects are flat, while CUBE and ROOF objects are thin enough for the robot to grasp.
The set of inferred types and predicates could be extended for more complex tasks (\eg object colour or orientation \cite{li2016learning}) but was beyond the scope of this work.

\subsection{Action Inference from Demonstration}
\label{sec:inference}
\begin{figure}
	\centering
	\includegraphics[width=0.9\linewidth]{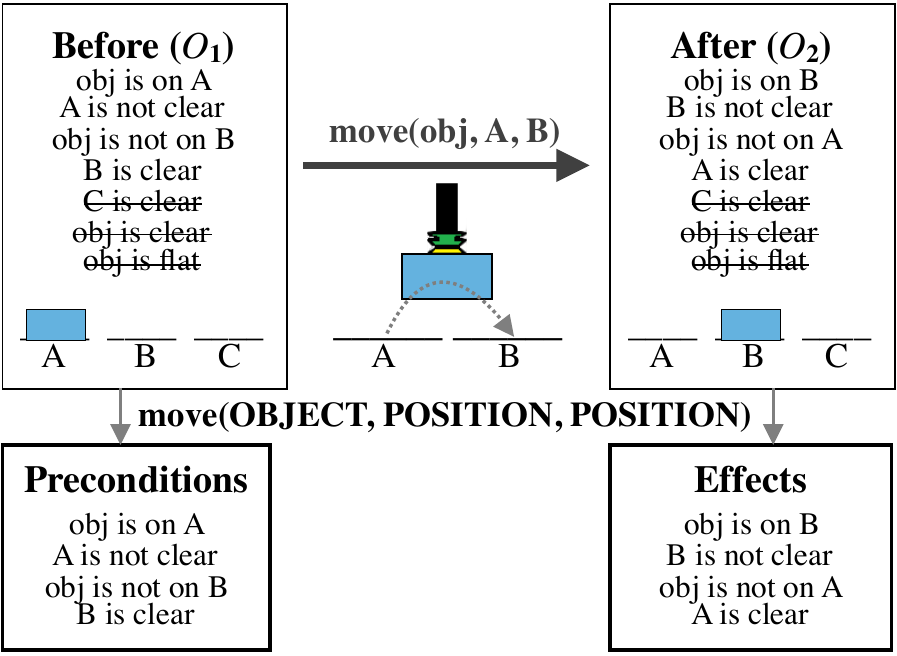}
	\caption{Example of a high-level action for moving an object from A to B. Conditions are inferred from the observed predicates before ($O_1$) and after ($O_2$) the demonstration \cite{liang2019d}.
	}
	\label{fig:action-model}
\end{figure}

Instead of manually defining action parameters, preconditions and effects, we accelerate the programming process by inferring them from the observed sensor data during the teaching phase.
Object types are inferred based on their detected bounding boxes.
Object positions are determined by the proximity of the object to given positions.
If the nearest position \emph{p} to the object \emph{o} is within a certain threshold $d$, then the predicates `\emph{o} is on \emph{p}' and `\emph{p} is not clear' are added to the detected world state.
To infer action conditions, the robot perceives the initial world state   before and after the action demonstration as seen in similar work for learning object manipulation tasks \cite{ahmadzadeh2015learning}.
Let $O_1 = \{\phi_1, \phi_2, ... \}$ be the set of predicates observed before the action demonstration and $O_2 = \{\psi_1, \psi_2, ... \}$ after.
The action inference is the heuristic deduction of predicates that have changed between $O_1$ and $O_2$, \ie
\begin{align*} \text{pre}(a) &= (O_1 - O_1 \cap O_2) = \{\phi_i | \phi_i \in O_1 \wedge \phi_i \notin O_2 \}, \\
\text{eff}(a) &= (O_2 - O_1 \cap O_2) = \{\psi_i | \psi_i \notin O_1 \wedge \psi_i \in O_2 \},
\end{align*}
where $\text{eff}(a)$ includes positive and negative effects (\fig{fig:action-model}).
A predicate $\phi$ has variables $\text{var}(\phi) = \{v_1, v_2, \dots\}$, where each $v_i$ has a type.
Therefore, action parameters are the set of variables that appear in either preconditions or effects, \ie
\begin{align*}
\text{param}(a) = \{v_i &| \hspace{0.3cm}\exists \phi \in \text{pre}(a) \text{ s.t. } v_i \in \text{var}(\phi)\\
&\lor \exists \psi \in \text{eff}(a) \text{ s.t. } v_i \in \text{var}(\psi) \}.
\end{align*}

Note that conditions could be learned from multiple demonstrations \cite{abdo2013learning,konidaris2018fromSkills}.
Our work argues for accelerating the teaching phase by learning from a single demonstration and letting the user act as the expert to correct wrongly inferred conditions.

\subsection{Action Generalisation}
\label{sec:generalisation}
The low-level action representation (\sect{sec:lowlevel}) generalises motion trajectories by re-calculating poses based on detected landmarks from the demonstrated to the new environment.
The high-level representation (\sect{sec:highlevel}) specifies when an action can be applied, therefore allows taught low-level motion trajectories to be reused for other objects (\eg use suction gripper for all objects, regardless of their type) or to be restricted for certain types (\eg only BASE objects).
By combining these two representation levels, taught actions can be generalised for new environments that are different to the demonstrated one, allowing the user to customise them for their specific use case.

\begin{figure}
	\includegraphics[width=\linewidth]{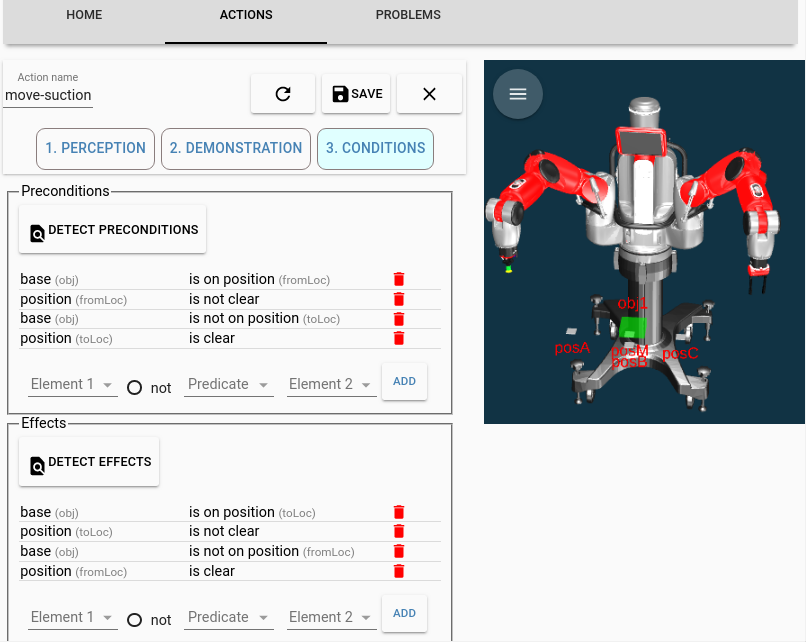}
	\caption{The iRoPro interface showing the action condition menu and an interactive visualisation of the Baxter robot and detected objects.}\label{fig:gui-action-3}%
\end{figure}

\begin{figure}
	\includegraphics[width=\linewidth]{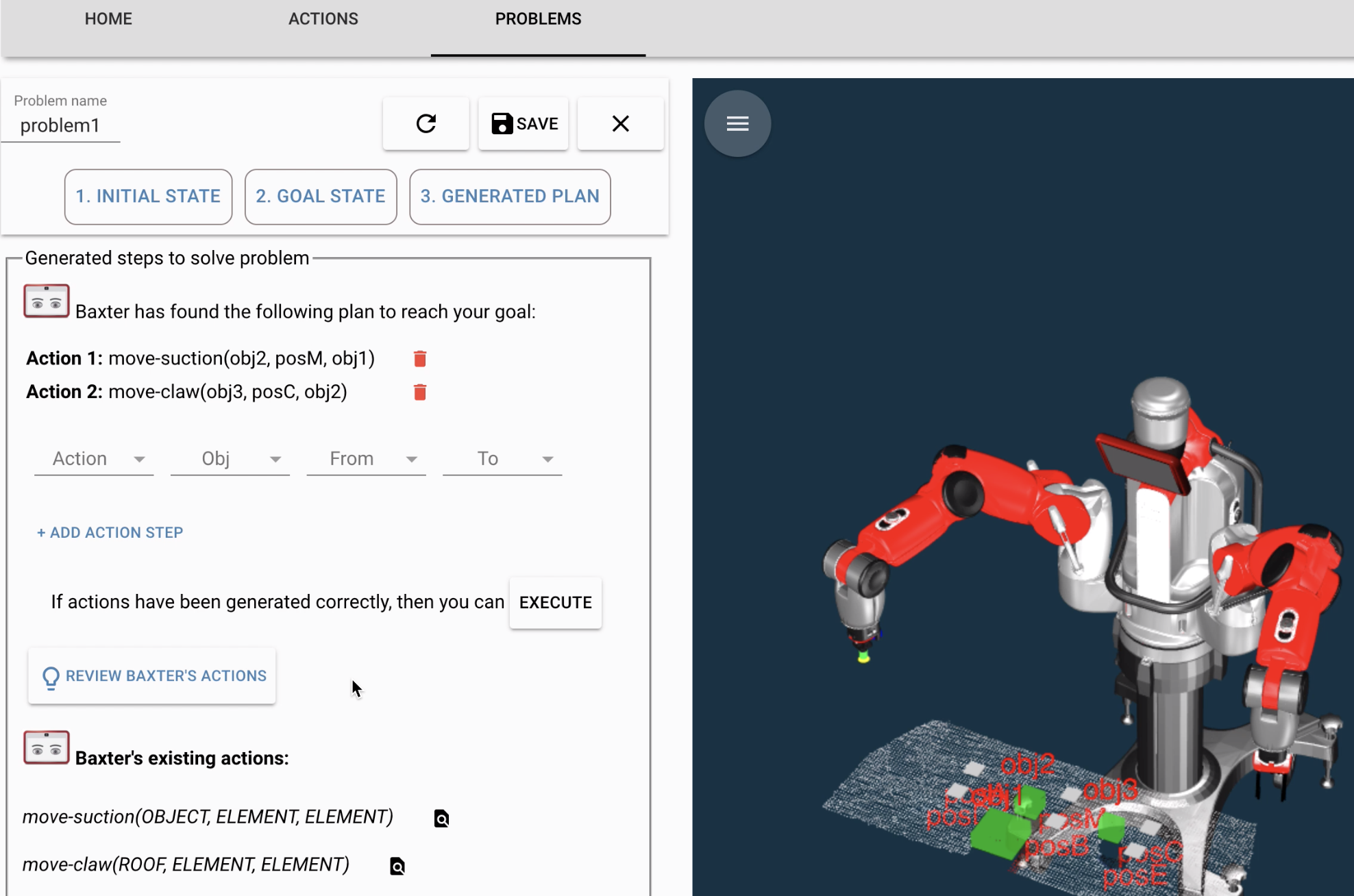}
	\caption{The iRoPro interface showing the problems menu and the action sequence generated by the task planner.}\label{fig:gui-problem}%
\end{figure}

\subsection{Interactive Robot Programming}
\label{sec:interactive}
The interacts with the GUI that visualises the robot and detected objects (\fig{fig:gui-action-3}).
During the programming process, the GUI allows them to create new actions, run the kinesthetic teaching by demonstration, modify inferred types and predicates and to create and solve new problems with the task planner.
The interactive robot programming cycle consists of creating and modifying actions and problems:
\paragraph{Actions.} New actions are taught by kinesthetically moving the robot's arms using kfPbD \cite{akgun2012keyframe}, where both low-level and high-level actions are learned and generalised (\sect{sec:generalisation}).
The user can have the robot re-execute the taught action immediately in order to validate it.
The user can modify the action properties if the inference was not correct.
To teach more actions, the user can either create a new one or copy a previously taught action and modify it.

\paragraph{Problems.} New planning problems can be generated if at least one action exists.
To create a problem, the robot first detects the existing landmarks and infers their types and initial states.
The user can modify them if the inference was not correct.
Then, the user enters predicates that describe the goal states to achieve.
The complete planning domain and problem are translated into PDDL and sent to the Fast-Forward planner \cite{hoffmann2001ff}.
If a solution is found that reaches the goal, it is displayed on the GUI for the user to verify and execute on the robot (\fig{fig:gui-problem}).
If no solution is found or if the generated plan is wrong, the user can open a debug menu to review actions.
It provides a summary of the entire planning domain with hints described in natural language to troubleshoot (\eg `make sure the action effects can achieve the goal states').
In our post-design study (\sect{sec:quanteval}) we found that this helped users understand how the system worked and why the generated plan was wrong.
Once the user modified actions, initial or goal states, they can relaunch the planner to see if a correct plan is generated.
For any subsequent tasks, the user can create a new problem or modify existing ones by re-detecting the objects.

\subsection{Plan Execution}
The generated plan is a sequence of actions with parameters that correspond to detected objects.
For each action, the sequence of end-effector poses are calculated relative to the landmarks that the action is being applied to (\sect{sec:lowlevel}).
To accelerate the execution, we only detect landmarks once at the start, then save their new positions in memory for quick reference, which we refer to as a \textit{mental model}.
After each action execution, the mental model is updated with the latest positions of the landmarks according to the action's effects (\eg obj moved from position A to B).
As this assumes that actions are always executed successfully, successful executions need to be checked separately but were beyond the scope of this work.
The mental model is also used as a workaround for our limited perception system as it does not detect stacked objects in their initial states (as discussed in \sect{sec:discussion}).

\section{Experimental Evaluation}\label{sec:exp-eval}
We first conducted two pre-design experiments to evaluate the usability of the proposed framework.
They consisted of qualitative user studies to respond to the following questions:
\begin{enumerate}
	\item[\textbf{Q1}] How do non-expert users adopt the Automated Planning language with its action model representation? (\sect{sec:Exp1})
	\item[\textbf{Q2}] Can users teach a robot action model for Automated Planning using the proposed framework? (\sect{sec:Exp2})
\end{enumerate}

In both experiments we particularly focused on elements to assess the user's understanding of action models such as defining their preconditions and effects.
Understanding this symbolic representation is a key requirement to use iRoPro.

Based on these results, we implemented iRoPro on the Baxter robot (\sect{sec:implementation}) and subsequently conducted post-design experiments to evaluate the working system with real end-users (\sect{sec:quanteval}).
Furthermore, we compared results obtained from pre-design with post-design experiments and validate the usability of our proposed framework (\sect{sec:exp-results}).
In the following sections we give a brief overview of the experimental setup, design, measurements and results.
Further details on all experiments can be found in previous work \cite{liang2017b,liang2019d}.

\subsection{Acceptance of Automated Planning concepts}\label{sec:Exp1}
In this experiment, we addressed the following question:
\begin{enumerate}
	\item[\textbf{Q1}] How do non-expert users adopt the Automated Planning language with its action model representation?
\end{enumerate}

Users (N=10) with little to no programming experience were introduced to the symbolic language and syntax with type structures used in Automated Planning.
Users were instructed to describe world state configurations to the robot.
The goal was to assess the user's adoption of the planning concepts (\eg object types, action models) and to verify that the symbolic planning language was appropriate for non-expert users.

\begin{figure}
	\centering
	\includegraphics[width=\linewidth]{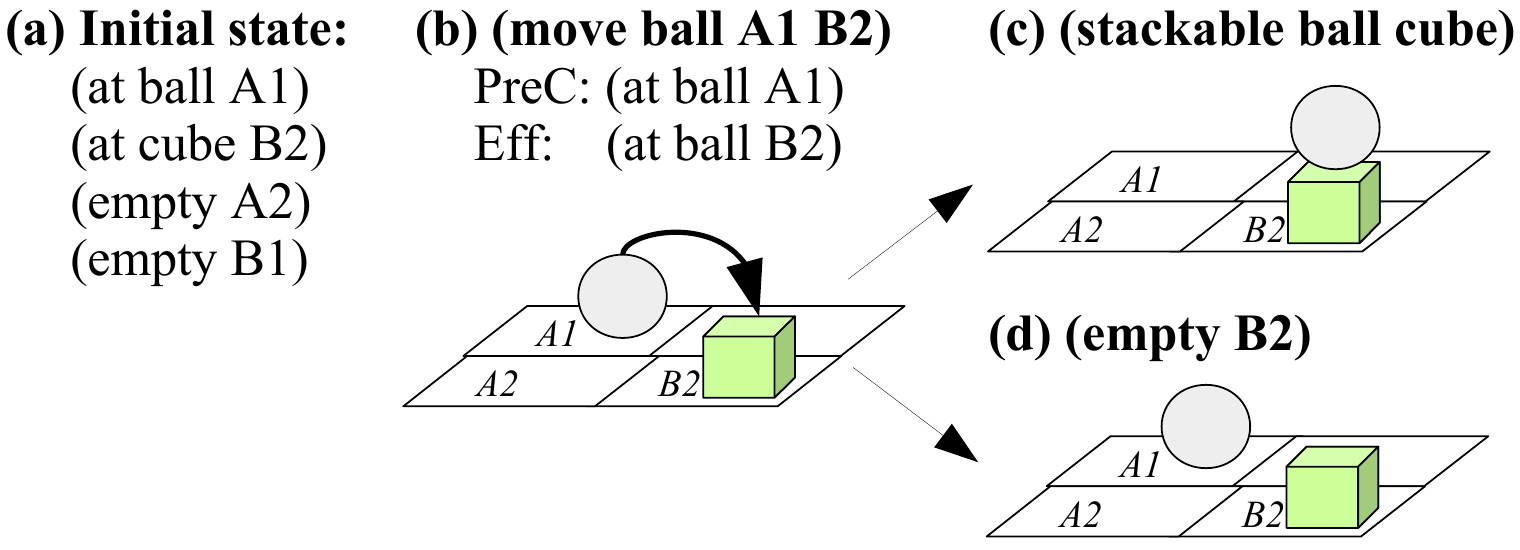}
	\caption{Users were instructed to provide a description of (a) the initial state of the world and (b) an initial move action model.
		Then they derived additional preconditions for moving the ball from position A1 to B2: (c) \textit{(stackable ball cube)}: the ball can be stacked onto the cube, and (d) \textit{(empty B2)}: if the ball cannot be stacked, the target position should be empty \cite{liang2017b}.}
	\label{fig:scenarios-exp1}
\end{figure}

\subsubsection{Experimental Design \& Measurements}
The experimental setup consisted of a 2x2 board (with positions A1, A2, B1, B2), 2 cubes, 1 ball, and 1 ball recipient in the form of a bowl.
The participants were given sheets with empty tables to complete for each task.
Each participant was allocated 1 hour, but the average duration of the experiment was 49 minutes.
Users were told that they needed to use a symbolic planning language to describe the state of the world and the semantic meaning of actions to the robot (\fig{fig:scenarios-exp1}).
At the course of the experiment, users were faced with three different scenarios of increasing complexity.
The participants' behaviour was observed by the experimenter and the experiment was recorded on camera.
We evaluated their capability to learn the presented planning language and apply it to different problem statements.
At the end, participants were given a questionnaire related to their experience and their understanding of the learned planning language and concepts.

\subsection{Acceptance of the Robot Programming Process} \label{sec:Exp2}
In this experiment, we addressed the following question:
\begin{enumerate}
	\item[\textbf{Q2}] Can users teach a robot action model for Automated Planning using the proposed framework?
\end{enumerate}
	Users (N=11)\footnote{All participants were different from the first experiment} with different programming experience were presented a simulated implementation of iRoPro and had to teach action models by kinesthetically manipulating a Baxter robot.
	Users were instructed to teach a primitive action by demonstration and assign preconditions and effects.
	The goal was to assess the framework's usability and the user's difficulties encountered during the programming process.
	At the end, participants were given a questionnaire related to their experience, their perceived understanding of the presented concepts and the usability of the framework.
	In the following sections we briefly outline the experimental setup, measurements and results of the experiment.

The experiments were conducted using a Baxter robot, mounted with a partial implementation of the framework.
The implemented functionalities included:
\begin{itemize}
	\item `learn new action': record the kinesthetic action demonstration,
	\item `find a coloured object': apply the recorded action to an object of the specified colour,
	\item `execute an action sequence': execute a sequence of previously taught actions.
\end{itemize}

We used the Wizard-of-Oz technique to simulate the remaining functionalities (\eg `infer action preconditions and effects', `generate solution using a planner').
Each participant was allocated 1 hour, but the average duration was 29.5 minutes.
The participants' behaviour was observed by the experimenter and the experiment was recorded on camera.

\begin{figure}
	\centering
	\includegraphics[width=\linewidth]{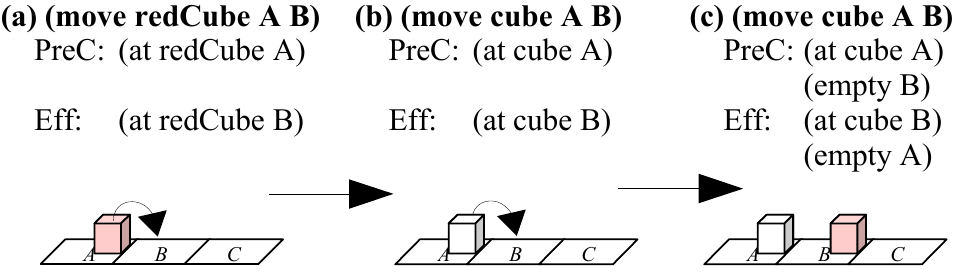}
	\caption{Continuous refinement of the move action model: (a) initial action model learned by demonstration, (b) action model for all cubes of any colour, (c) action model with an additional condition, if the target position is occupied and cubes can not be stacked \cite{liang2017b}.}
	\label{fig:scenarios-exp2}
\end{figure}

\subsubsection{Experimental Design \& Measurements}
The experiment scenario was set in a simulated assembly line, where objects of the same shape, but different colour arrived consecutively at a departure position.
Users had to teach the Baxter robot the action for moving an object from the departure position to an arrival position, where another maintenance task would be performed later.
Users first demonstrated the action on the robot by guiding its arm through the desired trajectory, then they were presented with an action model that the robot had `learned'.
At the course of the experiment, users were faced with two different scenarios, where they had to suggest logical conditions for the action model in order for the robot to apply them successfully (\fig{fig:scenarios-exp2}).
We evaluated the user's capability to improve action models and associate conditions when faced with different situations, and assessed the framework's overall usability.

\subsection{Pre-Design Experiments Findings}
In both experiments, we did not observe a significant difference in the performance between users with different programming experience.
In the first experiment, the majority (9 or 90\%) of the participants managed to describe the complete world state using the correct syntax.
All participants gave correct explanations for preconditions and effects of action models, and provided correct examples.
Figure \ref{fig:exp1vsexp2-results}a) shows the user responses to the questionnaire in the second experiment.
All 11 users were satisfied with the programming process and Baxter's ability to learn and reproduce the demonstrated move action.

The majority of the users had issues formulating the logical properties used for preconditions and effects.
In the first experiment (\sect{sec:Exp1}), users had difficulties formulating certain conditions in the planning language (\eg \texttt{(stackable ball cube)}), but stated equivalent ones (\eg \textit{`only place the ball, if it is stackable on the cube'}).
Similarly, in the second experiment (\sect{sec:Exp2}), users formulated missing preconditions (\eg \textit{`position B is empty'}) with other equivalent conditions (\eg \textit{`do not place the object on position B, if it is occupied'}).

Some of the users made wide assumptions about the robot's capabilities.
In the second experiment, when both arrival and departure positions were occupied, 5 (or 50\%) of the users expected Baxter to consider the occupied position, even though the condition was not mentioned in its action model.
This is a common problem in PbD solutions as there is a difference between the robot's intelligence and the one perceived by its teacher \cite{suay2012practical}.
This can be addressed by reproducing the learned action in a new context and verifying the robot's knowledge base, as we did throughout the experiment.

With these two qualitative experiments, we showed that the Automated Planning language and its main concepts can easily be learned by users without any programming background.
The action model representation, in terms of preconditions and effects, seems to be intuitive for non-expert users.
These initial experiments provided us with an initial idea of how the users might perceive the proposed robot programming framework.
We intentionally limited the set of Automated Planning concepts (\ie object types, predicates, and actions with preconditions and effects) that are necessary to use the framework to the bare minimum to assess the potential usability of such a framework.
Further experiments should test scalability, address more Automated Planning concepts (\eg object-type hierarchy, more predicates, planning problem definition and resolution) and potentially compare separate control groups (\eg experts vs non-experts) in less structured environments.

\subsection{Post-Design Experiment}
\label{sec:quanteval}
The second part of the evaluation was conducted using THEDRE \cite{mandran2017thedre}, a human experiment design method that combines qualitative and quantitative approaches to continuously improve and evaluate the developed system from the experimental ground.
The aim was to evaluate our approach with real end-users and we were also interested in the user's programming strategy for using the system.
We split participants into two control groups, with and without condition inference (\sect{sec:inference}) and evaluated user performance in terms of programming times for completing a set of benchmark tasks.
We set the following hypotheses for our experiments:
\begin{enumerate}
	\item[H1] Action creation: users can teach new low- and high-level actions by demonstration
	\item[H2] Problem solving: users can solve new problems by defining the goal states and executing the plan on Baxter
	\item[H3] Autonomous system navigation: users understand the system and can navigate and troubleshoot on their own
	\item[H4] Condition inference (CI) - Group 1 vs 2: users without CI will understand the system better
	\item[H5] Pre-study test (PT): users that score higher in the PT have shorter programming times
\end{enumerate}
In the following we will give a brief overview of the experiment setup, design, measurements and results.
Further details can be found in our previous work \cite{liang2019d}.

\begin{figure*}
	\begin{centering}
		\includegraphics[width=\linewidth]{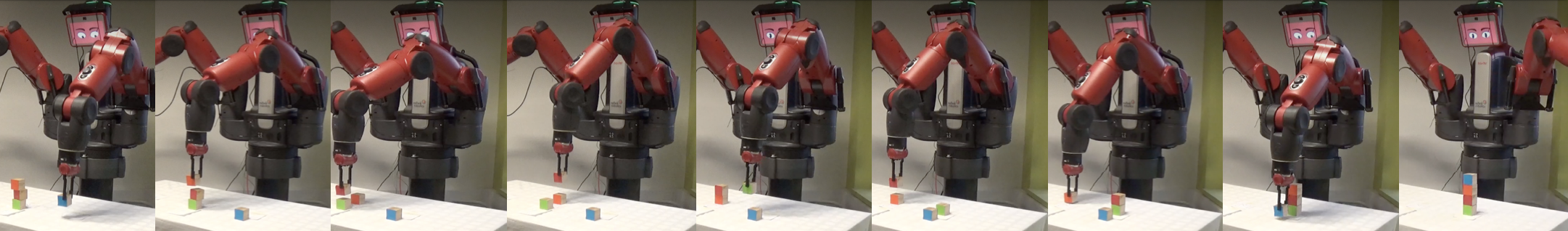}
		\caption{Snapshots of an example task execution for stacking CUBE objects with the claw gripper.}
		\label{fig:filmstrip}
	\end{centering}
\end{figure*}

\subsubsection{Participants}
The study was conducted with 21 participants (10M, 11F) in the range of 18-39 years (M=24.67, SD=6.1).
We recruited participants with different educational background and programming levels:
6 `CS' (either completed a degree in computer science or were currently pursuing one),
7 `non-CS' (have previously taken a programming course before),
and 8 `no experience' (only had experience with office productivity software).
Furthermore, 3 participants (in `CS') have programmed a robot before, out of which 1 had intermediate experience with symbolic planning languages while the remaining participants had no experience in either.
One participant in the category `non-CS' failed to complete the majority of tasks and was excluded from the result evaluation.
The two control groups included equal number of participants in all three categories.

\subsubsection{Protocol}
Users were first given a brief introduction to task planning concepts, the Baxter robot and the experimental set up (\fig{fig:dispositif}).
They were then asked to complete a pre-study test to capture the participant's understanding of the presented concepts.
Users were given 8 tasks to complete (Table \ref{table:userstudytasks}), where the first two were practice tasks to introduce them to the system.
The tasks were designed to address different aspects to familiarise them with the system:
create new actions (Task 6), modify parameter types (Tasks 4\&7), modify action conditions (Tasks 3,5,8).
For each task they needed to create a new problem, define the goal states, and launch the planner to generate an action sequence.
When the generated plan was correct, they were executed on the robot (\fig{fig:filmstrip}).
Otherwise, the user had to modify the existing input until the plan was correctly generated.
Tasks 6-8 were similar to the previous tasks (1-5) but use both robot grippers.

\subsubsection{Metrics}
We captured the following data during the experiments:
\begin{enumerate}
	\item \textbf{Qualitative data:} video recording of the experiment, observations during the experimental protocol.
	\item \textbf{Quantitative data:} task duration and UI activity log, pre-study test, post-study survey.
\end{enumerate}

The pre-study test included seven questions related to their understanding of the concepts presented at the start of the experiment, \eg syntax (`If move(CUBE) describes a move action, tick all statements that are true.'), logical reasoning
(`Which two conditions can never be true at the same time?'), and other concepts (`Tick all predicates that are required as preconditions for the given action').
The questions were multiple choice and the highest achievable score was 7.

In the post-study survey we used the System Usability Scale (SUS) \cite{brooke2013sus} where participants had to give a rating on a 5-Point Likert scale ranging from `Strongly agree' to `Strongly disagree'.
It enabled us to measure the perceived usability of the system with a small sample of users.
As a benchmark, we compared overall responses to the second user study, where users were simulated a similar robot programming experience using the Wizard-of-Oz technique but had no direct interaction with a working system (\sect{sec:Exp2}).
Finally, participants were asked which aspects they found most useful, most difficult, and which they liked the best and the least.

\begin{table}
	\centering
	\caption{Benchmark tasks for the user study where the first two tasks were used to introduce participants to the system \cite{liang2019d}.}
	\label{table:userstudytasks}
	\begin{center}
		\begin{tabular}{l@{\hskip2.0pt}l}
\hline\noalign{\smallskip}
			\textbf{\# Task description} & \textbf{Main solution} \\
\noalign{\smallskip}\hline\noalign{\smallskip}
			(1) move BASE object (suction grip) & create new action (+demo) \\
			(2) move BASE object to any position & create new problem \\
			3 swap two BASE objects & add condition (`is clear') \\
			4 stack CUBE on BASE & modify types (`OBJECT')\\
			5 do not stack CUBE on ROOF & add condition (`is stackable')\\
			6 move ROOF object (claw grip) & create new action (+demo) \\
			7 stack ROOF on a CUBE & modify types (`ELEMENT') \\
			8 build a house (BASE, CUBE, ROOF) & navigate autonomously \\
\noalign{\smallskip}\hline
		\end{tabular}
	\end{center}
\end{table}

\begin{figure*}
	\includegraphics[width=0.98\linewidth]{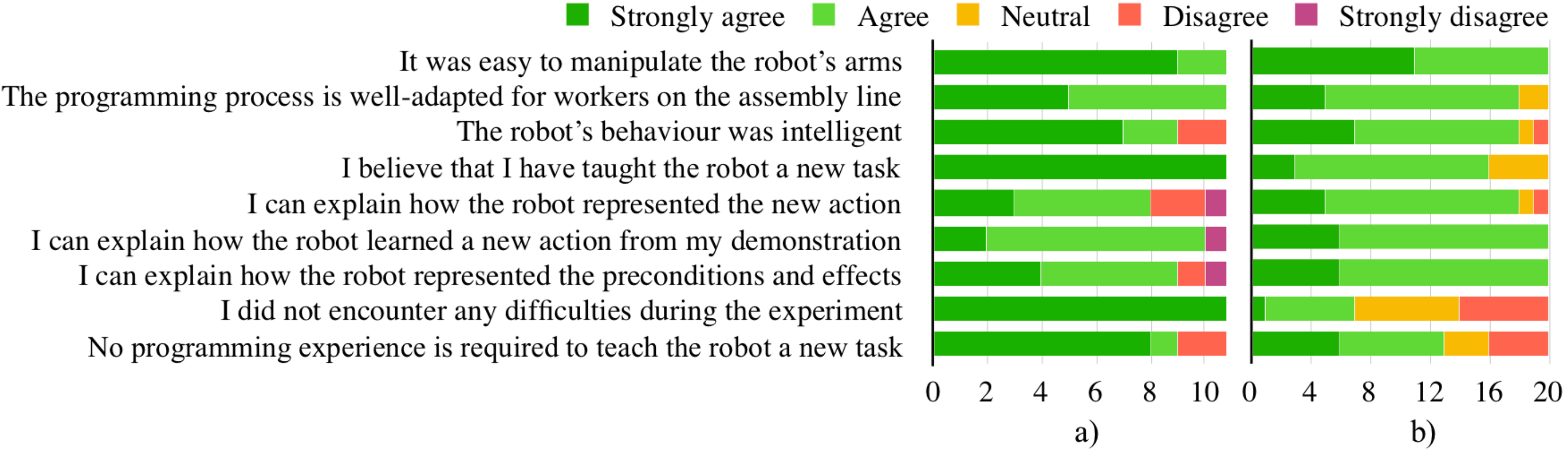}
	\caption{User responses from the post-study survey comparing a) the pre-design study (\sect{sec:Exp2}) with b) the post-design study (\sect{sec:quanteval}).}
	\label{fig:exp1vsexp2-results}
\end{figure*}

\subsection{Results}\label{sec:exp-results}
20 participants completed all tasks, while one `non-CS' user failed to complete the majority of tasks and did not seem to understand the presented concepts.
This participant was excluded in the results presented below (Table \ref{table:performance}):

\subsubsection*{\textbf{H1)-H3) User performance}}
Users took between 22-60 minutes to complete the main tasks (3-8), with an average of 41.2 minutes.
`non-CS' users completed the tasks the fastest, followed by users with no programming experience.
`CS' users took on average longer as they were often interested in testing the system's functionalities that were beyond the given tasks.

Users initially had problems with different concepts that were presented at the start of the study, in particular they confused action parameters, preconditions and goal states.
For example, in Task 3, 6 (or 30\%) users tried to add intermediate action steps to achieve the goal state, instead of simply letting the planner generate the solution.
In Task 4, 14 (or 70\%) wanted to create a new action, even though they could reuse the existing action by modifying the parameter types.
However, by Task 6, all users were able to use the system autonomously to create new actions and problems and navigated the system with little to no guidance.
By the end of the experiment, users programmed two manipulation actions (one for each gripper) that were reused to complete all 8 benchmark tasks.

\subsubsection*{\textbf{H4) Condition inference (CI)}}
We noticed a discrepancy in the programming strategies between the two control groups (Group 1 with CI vs. Group 2 without CI).
Participants in Group 1 had the tendency to leave the inferred conditions unmodified without adding conditions that would improve the action's generalisability to different use cases.
As participants in Group 2 had to add action conditions manually, they considered all predicates they deemed necessary for the action and added additional ones that were required for later tasks.
Thus, Group 2 took on average longer to complete tasks where a new action had to be created (Tasks 1\&6), but was faster than Group 1 for subsequent tasks, where conditions had to be modified (Tasks 3,5,7).
Overall both groups had similar completion times for all tasks.

\subsubsection*{\textbf{H5) Pre-study test}}
As expected, participants who demonstrated a better understanding of the introduced concepts in the pre-study test completed the main tasks (Tasks 3-8) faster on average (p-value$<0.05$).
Users scored between 4.3-6.93 out of 7 points.
`non-CS' users scored above average points and completed the fastest.
As an outlier we observed that the fastest participant scored only 4.7, but easily learned how to use the system and completed the tasks in 22 minutes.
Even though Group 1 performed slightly better in the pre-study test than Group 2, both completion times were on average similar.

\begin{table}
	\centering
	\caption{User performance comparing task completion times with pre-study test scores \cite{liang2019d}.}
	\label{table:performance}
	\begin{center}
		\begin{tabular}{r|rr|rr}
			& \multicolumn{2}{c}{\textbf{Main tasks (in min)}} & \multicolumn{2}{c}{\textbf{Pre-score (out of 7)}} \\
			& AVG                & STD                & AVG                   & STD                   \\ \hline
			no experience  & 43.6               & 5.37               & 5.8                   & 0.95                  \\
			non-CS  & 36.6               & 7.46               & 6.2                   & 0.55                  \\
			CS      & 43.8               & 14.13              & 5.3                   & 1.11                 \\ \hline
			\textbf{Overall} & \textbf{41.2}              & \textbf{9.08}               & \textbf{5.8}                   & \textbf{0.91}                  \\ \hline
			Group 1 & 41.0               & 7.89               & 6.1                   & 0.77                  \\
			Group 2 & 41.4               & 10.56              & 5.5                   & 0.98
		\end{tabular}
	\end{center}
\end{table}

\subsubsection*{\textbf{System usability and learnability}}
There are several ways to interpret the System Usability Scale (SUS) scores \cite{brooke2013sus} obtained from the post-study survey.
Using Bangor et al.'s categories \cite{bangor2008suseval}, 14 (70\%) users ranked iRoPro as `acceptable', 6 (30\%) rated it `marginally acceptable', and no one ranked it `not acceptable'.
Correlating this with the Net Promoter Score \cite{nps}, this corresponds to 10 (50\%) participants being `promoters' (most likely to recommend the system), 5 (25\%) `passive', and 5 (25\%) `detractors' (likely to discourage).
Overall, iRoPro was rated with a good system usability and learnability.

9 (45\%) users stated `generate solutions to defined goals automatically' as the most useful feature, followed by `robot learns action from my demonstration' (4 or 20\%) -- two main aspects of our approach.
4 (20\%) stated the most difficult part as `finding out why Baxter didn't solve a problem correctly', similarly 8 (40\%) stated difficulties related to `understanding predicates and defining conditions'.
11 (55\%) disliked `assigning action conditions' the most, while the rest stated different aspects.
A common feedback was `it takes time to understand how the system works at the start'.
The most liked parts were `executing the generated plan' (8 or 40\%) and `demonstrating an action on Baxter' (7 or 35\%).

\subsection{Pre- vs Post-Design Experiments}
We compare post-study questionnaire responses between pre- and post-design experiments (\fig{fig:exp1vsexp2-results}).
In the first experiment (\sect{sec:Exp2}), users had no direct interaction with the robot programming system as it was simulated using the Wizard-of-Oz technique, while in the last experiment (\sect{sec:quanteval}), users programmed the robot using the end-to-end system implementation.
The main differences were noted regarding difficulties encountered during the experiment:
In the first study 11 (or 100\%) users agreed that they encountered no difficulties, while in the last study only 7 (or 35\%) users stated the same.
However, all users in the last study claimed to have a good understanding of the action representation and how the robot learned new actions from their demonstrations, while an average of 2 (18\%) disagreed in the first study.
Both differences can be explained by the fact that in the last study, users had to use an end-to-end system to program the robot, while in the first study, users had no direct interaction with a working system.
Even though users encountered more difficulties in the last study, they got a better understanding of the functionalities due to getting hands-on experience.
This also correlates with negative responses in our survey to the question if `no programming experience was required' where 13/20 (65\%) agreed and 4 disagreed.
Overall, both user studies received positive responses.

\subsection{Continuous Improvement of the System}
\begin{figure}
	\centering
	\includegraphics[width=\linewidth]{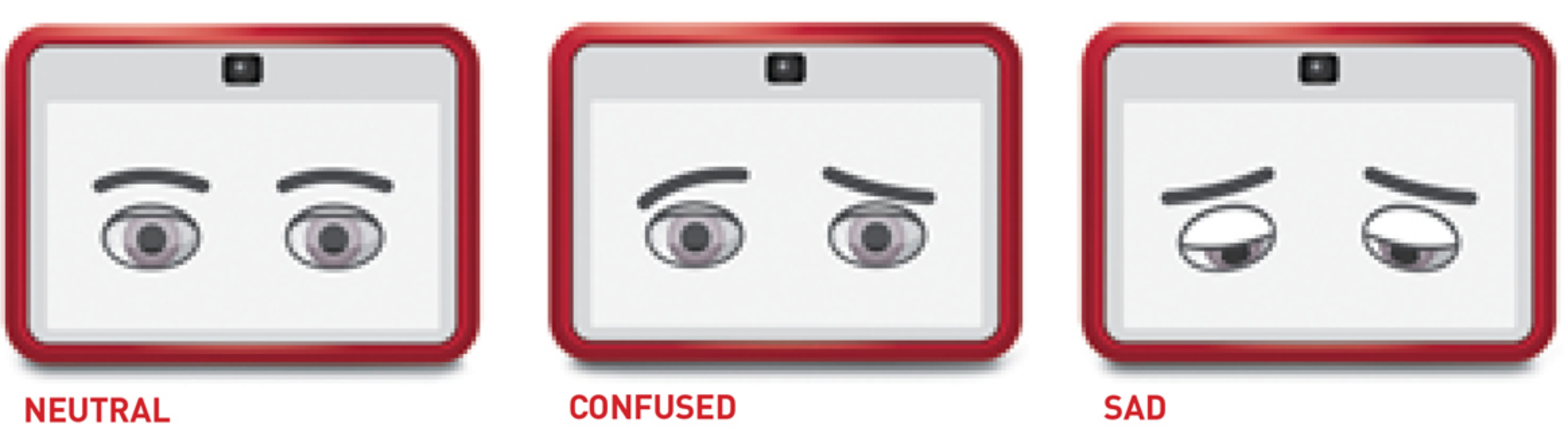}
	\caption{Baxter icons used for the graphical interface \cite{baxteremojis}.}
	\label{fig:baxter-icons}
\end{figure}

The system underwent four phases of improvement, allowing us to refine the system functionalities, GUI and user instruction methods:
\begin{enumerate}
	\item {\textbf{Initial prototype:}
		Based on the feedback received on our initial prototype by domain experts, 
		we changed the flow for introducing the system to novice users as it included a lot of new information (\eg PbD and Automated planning concepts) that they were not familiar with.
	}
	\item {\textbf{Pre-tests with 3 users:}
		We ran pre-tests with 3 users who have never seen the system before and further improved the experiment flow (\eg create an action for one object at a time).
		We also made the user interface more friendly to include Baxter icons (\fig{fig:baxter-icons}) and eyes that followed the robot's moving joint \cite{baxtereyes}) on the robot's screen so that it seemed more human-like.
		We noticed that troubleshooting incorrect actions seemed difficult for all users, so we included an option to review actions and goals which provides a summary of the created input and hints to guide the debugging steps.}
	\item {\textbf{First experimental tests with 5 users with condition inference:}
		After running the experiments with 5 users, we noticed that the majority did not modify the inferred action conditions.
		This raised an interesting HRI question of whether users who were given automatically generated conditions would `blindly trust' them.
		Hence, we decided to create two experiment groups: with and without condition inference, where the first 5 participants belonged to the former.
		Participants in the latter group would need to manually enter all conditions via the interface. }
	\item {\textbf{Final experimental tests with 16 users:}
		The remaining users were divided into the two control groups so that we had an equal number of participants in both control groups, while also maintaining an even distribution of programming levels.
		The results of the 21 participants in the experimental tests were used as the final evaluation of the system as discussed in the previous sections.
	}
\end{enumerate}

\section{Discussion}
\label{sec:discussion}
We demonstrated that iRoPro can be used to generalise primitive actions to a range of complex manipulation tasks and that it is easy to learn for users with or without programming experience.
In our system evaluation we could have programmed other actions, such as turning or pushing for packaging tasks \cite{liang2018c}.
As the purpose of our evaluation was to show the generalisability of primitive actions with the use of a task planner, we decided to stick to pick-and-place actions with two different grippers.
Limitations and interesting extensions of our work are as follows:
\begin{enumerate}
	\item Our object perception is limited as it does not detect objects that are too close together (\eg stacked objects).
	An improved perception system would allow the detection of initial states with stacked objects, automatically detecting goal states, or verifying action executions.
	\item Due to the different grippers, we did not program actions that use both arms (\eg carrying a tray). A possible extension would be to include a better motion and task planning system in order to allow executing both arms simultaneously while avoiding self-collision.
	\item We only included a minimal set of predicates (\sect{sec:highlevel}) that we deemed intuitive and useful for object manipulation tasks.
	It could be interesting to include and learn predicates to capture more complex domains such as object orientation \cite{li2016learning}.
	\item iRoPro can easily be adapted to different kinds of robots: the source code is available online (\sect{sec:implementation}) and developed in ROS, which is a popular framework facilitating the interoperability of a wide variety of robotic platforms. Moreover, the PDDL language used to encode planning domains is platform-independent, and a standard used to benchmark task planners. However, we need to push our standardization effort further by bridging the gap between PDDL and IEEE Standard Ontologies for Robotics and Automation \cite{8172300}.
\end{enumerate}

\section{Conclusion}
\label{sec:conclusion}
We presented iRoPro, an interactive Robot Programming system that allows simultaneous teaching of low- and high-level actions by demonstration.
The robot reuses the actions with a task planner to generate solutions to tasks that go beyond the demonstrated action.
We conducted pre-design experiments to evaluate the feasibility of the proposed framework and involved concepts in PbD and Automated Planning.
Then, we implemented the system on a Baxter robot and showed its generalisability on six benchmark tasks by teaching a minimal set of primitive actions that were reused for all tasks.
We demonstrated its usability with a user study where participants with diverse educational backgrounds and programming levels learned how to use the system in less than an hour.
Furthermore, we compared user responses from our pre-design experiments with our post-design evaluation and investigate discrepancies.
Both user performance and feedback confirmed iRoPro's usability, with the majority ranking it as `acceptable' and half being promoters.
Overall, we demonstrated that our approach allows users with any programming level to efficiently teach robots new actions that can be reused for complex tasks.
Thus, iRoPro enables end-users to program robots from scratch, without writing code, therefore maximising the generalisability of taught actions with minimum programming effort.

\section{Future Work}
\label{sec:futurework}
Future work will focus on exploring more challenging domains to extend the system to other platforms by including a wider range of predicates and probabilistic techniques to improve the condition inference.
The next step is to focus on less controlled environments involving factory workers who may ultimately use this technology as well as mobile robots that move around between workspaces.
This will require more complex planning domains to consider planning between robots and humans, human-robot collaborative tasks as well as solutions for multi-modal communication involving natural cues \cite{pais2013assessing}.



\begin{acknowledgements}
The authors would like to thank Nadine Mandran, research engineer at the Laboratoire d'Informatique de Grenoble, for her support and guidance during for the experimental studies.
\end{acknowledgements}

%
%

\bibliographystyle{spmpsci}      
\bibliography{main}


\end{document}